\documentclass{egpubl}
\usepackage{pg2022}

\SpecialIssuePaper         

\usepackage[T1]{fontenc}
\usepackage{dfadobe}  

\usepackage{cite}  
\BibtexOrBiblatex
\electronicVersion
\PrintedOrElectronic
\ifpdf \usepackage[pdftex]{graphicx} \pdfcompresslevel=9
\else \usepackage[dvips]{graphicx} \fi

\usepackage{egweblnk}

\usepackage{color}
\usepackage{multirow}
\usepackage{booktabs} 
\usepackage{bm}
\usepackage{amsmath, amsfonts}

\title[Generative Deformable Radiance Fields for Disentangled Image Synthesis of Topology-Varying Objects]%
      {Generative Deformable Radiance Fields for Disentangled Image Synthesis of Topology-Varying Objects}

\author[Z. W\&Y. D\&J. Y\&J. Y\&X. T]
{\parbox{\textwidth}{\centering Ziyu Wang\thanks{Work done when ZW and YD were interns at MSRA. \\Project page: \textcolor{blue}{https://ziyuwang98.github.io/GDRF/}}$^{1}$\orcid{0000-0002-4697-5183}, Yu Deng\footnotemark[1]$^{2,3}$\orcid{0000-0001-7241-8519}, Jiaolong Yang$^{3}$\orcid{0000-0002-7314-6567}, Jingyi Yu$^{1}$\orcid{0000-0001-9198-6853} and Xin Tong$^{3}$\orcid{0000-0001-8788-2453}
        }
        \\
{\parbox{\textwidth}{\centering $^1$ ShanghaiTech University, \
         $^2$ Tsinghua University, \
         $^3$ Microsoft Research Asia
       }
}
}

\begin{document}

\teaser{
 \includegraphics[width=1.0\linewidth]{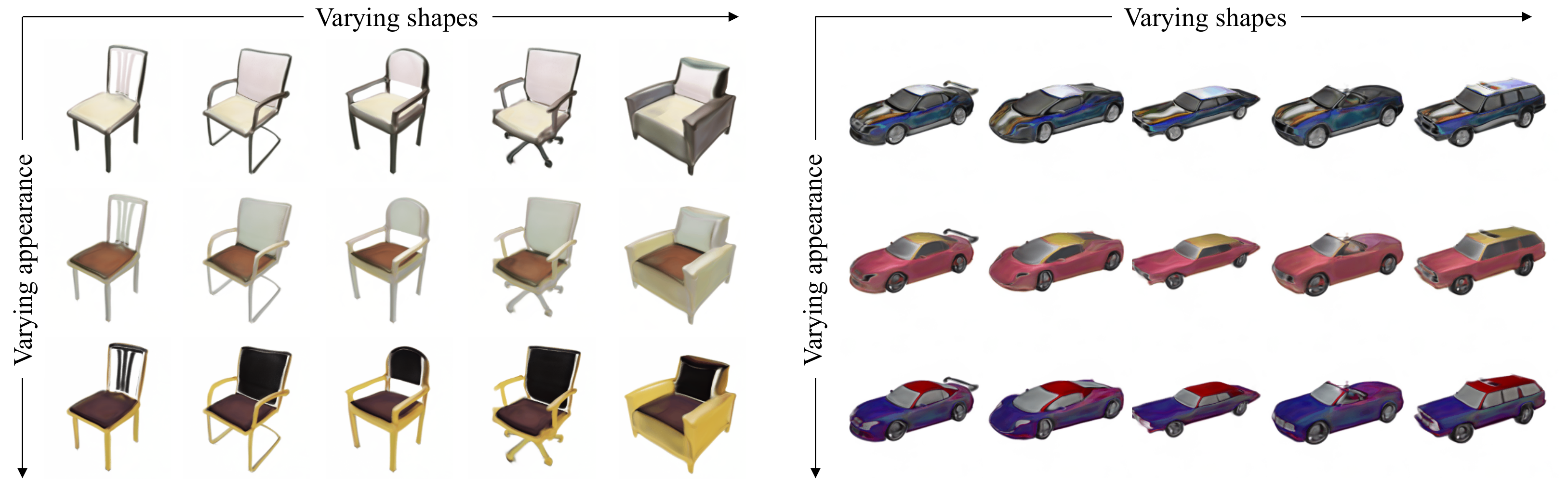}
 \centering
  \caption{Image generated by our method with disentangled shape and appearance synthesis. Texture of the corresponding semantic components can be correctly transferred between different shapes with topology variations. \textbf{See the accompanying video for multiview results.}}
\label{fig:teaser}
\vspace{10pt}
}

\maketitle
\begin{abstract}

3D-aware generative models have demonstrated their superb performance to generate 3D neural radiance fields (NeRF) from a collection of monocular 2D images even for topology-varying object categories. However, these methods still lack the capability to separately control the shape and appearance of the objects in the generated radiance fields. In this paper, we propose a generative model for synthesizing radiance fields of topology-varying objects with disentangled shape and appearance variations. Our method generates deformable radiance fields, which builds the dense correspondence between the density fields of the objects and encodes their appearances in a shared template field. Our disentanglement is achieved in an unsupervised manner without introducing extra labels to previous 3D-aware GAN training. We also develop an effective image inversion scheme for reconstructing the radiance field of an object in a real monocular image and manipulating its shape and appearance. Experiments show that our method can successfully learn the generative model from unstructured monocular images and well disentangle the shape and appearance for objects (e.g., chairs) with large topological variance. The model trained on synthetic data can faithfully reconstruct the real object in a given single image and achieve high-quality texture and shape editing results.


\end{abstract}  
\section{Introduction}

3D-aware GANs~\cite{schwarz2020graf,chan2021pi,wang2021clip} are capable of learning neural radiance field (NeRF)~\cite{mildenhall2020nerf} of an object category with topology variations, given only a collection of monocular 2D images as supervision. With the advantages brought by the NeRF representation, they can easily generate images at arbitrary viewpoints and maintain multiview consistency when varying camera views, which greatly improves the prospects of GAN-based methods in applications like VR\&AR. 

However, a shortcoming of these methods is  lack of controllability over factors beyond viewpoints, especially a disentangled control over shape and appearance. Such a limitation is not only seen in NeRF-based 3D-aware GANs, but also commonplace in existing methods that utilize NeRF representation for object modeling~\cite{liu2021editing,jang2021codenerf}. This reduces their practical values when applied to image editing tasks such as texture transfer and shape manipulation. Although some methods~\cite{schwarz2020graf,wang2021clip} address this problem by designing separate latent codes for shape and appearance and send them to different layers of the NeRF generator, they still cannot guarantee independent changes because the predicted shape and color still share some common hidden features within the network.

Another line of  works~\cite{blanz1999morphable,loper2015smpl,groueix2018papier,jiang2020shapeflow,zheng2020dit,deng2021deformed} for shape modeling have shown that deriving dense correspondence via deformations between object surfaces or their implicit fields can well address the disentanglement between appearance and shape. By learning reasonable correspondence between different shape surfaces, textures can be correctly transferred between corresponding semantic regions or structures, thus achieving high-quality disentanglement between shape and appearance control. However, most of these methods are trained with a 3D auto-decoding paradigm where ground truth 3D shapes with adequate geometry clues can be utilized for reasonable deformation learning. It is unclear how to model the dense correspondence between radiance fields and whether these methods can be adapted to the challenging 3D-aware GAN scenario without direct 3D supervision.

In this paper, we propose a novel generative deformable radiance fields that introduces 3D deformation into adversarial learning of NeRF-based GANs using only 2D images. With our method, disentangled control over shape and appearance can be achieved for image synthesis of objects with topology changes. The key idea is to encode the shapes of generated instances via deformation of a template radiance field shared across the category, and model the appearance within the shared template radiance field. 
Nevertheless, previous methods~\cite{groueix2018papier,zheng2020dit,deng2021deformed} have shown that pure deformation cannot handle correspondences for objects with topology difference and hence leads to undesired results for shape and texture editing. Therefore, we further add a correction field on top of the deformed radiance fields to handle structure changes and allow reasonable dense correspondence reasoning. 
To incorporate these modules into the 3D-aware GAN framework, we develop a set of carefully-designed loss functions to effectively learn the deformed radiance field with disentangled shape and appearance latent vector from unstructured monocular 2D images. With the learned generative deformable radiance fields, we further design an effective inversion initialization method to achieve high-quality image embedding and editing results.


We evaluate the proposed method on multiple datasets of topology-varying objects, including Photoshape~\cite{photoshape2018}, ShapeNet~\cite{wu20153d}, and CARLA~\cite{dosovitskiy2017carla,schwarz2020graf}. We show that our method can not only generate 3D-consistent multiview images of these objects, but can also control their shapes and appearance in a disentangled manner, which greatly improves the controllability for NeRF-based image synthesis. In addition, we demonstrate the ability of our proposed image inversion scheme which can faithfully reconstruct real objects in an image and achieves high-fidelity texture editing and novel-view synthesis, even though we only trained our generative models on synthetic data. This reveals the strong potentials of our method for realistic virtual content creations and manipulations in future applications.


\section{Related Work}
\noindent\textbf{Neural scene representations.} Neural scene representation methods have witnessed significant progress in recent years, from the early black-box CNNs~\cite{kulkarni2015deep,dosovitskiy2016learning, tatarchenko2016multi,isola2017image,sitzmann2019deepvoxels} to the latest neural implicit fields~\cite{park2019deepsdf,mescheder2019occupancy,sitzmann2019scene,sitzmann2020implicit,mildenhall2020nerf,wang2021neus} and hybrid representations~\cite{yu2021plenoxels,peng2020convolutional,chan2022efficient,peng2021neural,lombardi2021mixture,muller2022instant}. A representative method among them is neural radiance field (NeRF)~\cite{mildenhall2020nerf}, which leverages an MLP to predict color and density of each point in a 3D volumetric space for scene representation. NeRF and its follow-up works~\cite{barron2021mip,barron2022mip,park2020deformable,tancik2022block,schwarz2020graf,chan2021pi,liu2021editing,jang2021codenerf} have demonstrated their strong capabilities in learning novel view synthesis of 3D scenes given only 2D images for training. However, they do not address explicit control over other properties of the scene such as shape and appearance. Several methods utilize two individual latent codes and send them to different layers of the radiance field network for certain degree of disentanglement between shape and appearance. They achieve color transfer at a global level but cannot handle detailed textures. Some recent methods adopt deformation fields for better shape and appearance disentanglement, but they either target at single scene reconstruction and editing~\cite{xiang2021neutex} or object categories without large topology changes~\cite{xie2021fig,tewari2022disentangled3d}. We propose a new method to model topology-varying object categories and independently control the shape and appearance within a 3D-aware generative modelling scheme.

\begin{figure*}[t]
    \centering
    \includegraphics[width=\linewidth]{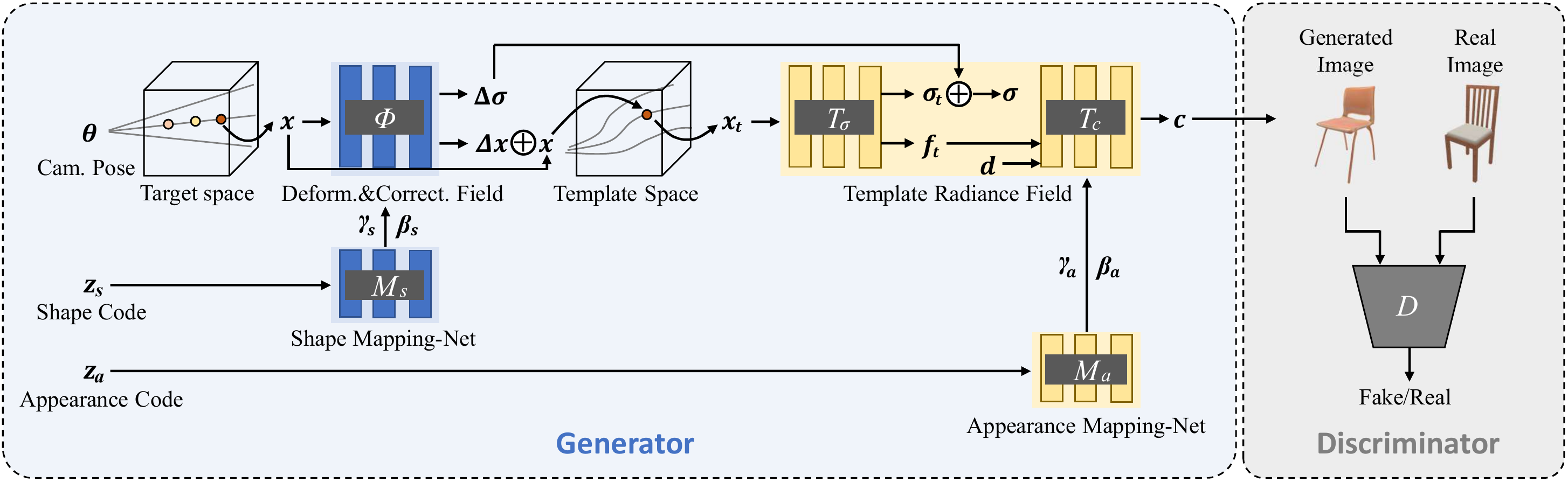}
    \vspace{-12pt}
    \caption{Overview of our generative deformable radiance field. It consists of a template radiance field conditioned on a appearance code $z_a$, and a deformation field and a correction field conditioned on a shape code $z_s$. A discriminator is introduced for adversarial learning between generated images and real ones. All networks are train end-to-end on a collection of color images.} 
    \label{fig:frame}
    \vspace{-5pt}
\end{figure*}

\noindent\textbf{3D-aware image synthesis.~} Recent 3D-aware GANs \cite{nguyen2019hologan,schwarz2020graf,chan2021pi,niemeyer2021giraffe,devries2021unconstrained,deng2021gram,chan2022efficient} are capable of learning multiview image generation of an object category from a monocular image set, which greatly simplifies the training configuration and demonstrates their potentials for large-scale image synthesis applications. Among them, images synthesized by NeRF-based generators~\cite{schwarz2020graf,chan2021pi,devries2021unconstrained} have demonstrated more strict 3D-consistency when varying camera views, which is an indispensable feature for the multiview image generation task. However, they still cannot achieve satisfactory control over shape and appearance. Perhaps the most relevant method to ours is in a very recent work of Tewari et al.~\cite{tewari2022disentangled3d} that is concurrent to ours. In this method, a deformation network is used to separate shape variations from a canonical radiance field that handles appearance changes, and an inverse deformation field is applied to enforce invertible mapping for reasonable deformation learning. However, this method does not deal with objects that have structure variance. In this paper, we incorporate both deformation field and correction field into NeRF-based 3D-aware GAN to handle more complex objects such as chairs.

\noindent\textbf{Shape modeling via deformations.~} Deformation-based modeling of a category of 3D shapes is a longstanding task in the literature and still an active topic to date. The deformations build dense correspondence between different shapes thus appearance of them can be easily defined on a shared template to achieve strong disentanglement with the geometry.
Early works~\cite{blanz1999morphable,vlasic2006face,anguelov2005scape,paysan20093d,loper2015smpl,zuffi20173d} build linear subspace to characterize deformations between registered meshes. They are restricted to object categories with consistent topology and structures such as face and human body. Later works~\cite{groueix2018papier,yang2018foldingnet} handle more complex objects by learning nonlinear mesh deformations from a large number of 3D shapes via deep neural network. Recently, some methods~\cite{deng2021deformed,zheng2020dit,zheng2022imface,yenamandra2021i3dmm} combine deformations with implicit shape representations. Among them, \cite{deng2021deformed} introduces an extra correction field to deal with structure discrepancies between shapes that are difficult to handle by continuous deformations. It shows promising results for detailed texture transfer between 3D shapes. Different from the aforementioned methods, we learn deformations and corrections in 3D space given only 2D images as training data, which is significantly more challenging than their 3D auto-decoding setup with ground truth 3D meshes for supervision.


\noindent\textbf{Disentangled image synthesis and manipulations.~} A large volume of methods focus on disentangled control over multiple attributes of 2D image contents.
Some methods~\cite{chen2016infogan,higgins2016beta} propose to learn disentangled representation by 2D CNNs in a fully unsupervised manner but often cannot guarantee disentanglement between semantically meaningful attributes. More of existing works~\cite{dosovitskiy2016learning,tran2017disentangled,pumarola2018ganimation,shen2018faceid,park2019semantic,deng2020disentangled,abdal2021styleflow} leverage certain kinds of attribute labels as guidance to achieve controllable image synthesis and editing of desired properties. Nevertheless, they show inferior results for 3D pose disentanglement compared to methods based on state-of-the-art 3D representations. Our method utilizes the NeRF representation thus can not only achieve disentangled image manipulation of shape and appearance but also synthesize their novel views with high 3D consistency.


\section{Overview}
Given a monocular image collection of an object category with topology variations, we aim to learn an image generator $G$ that can synthesize multiview images of virtual instances and control their shape and appearance attributes independently. Specifically, given a shape code $\bm z_{s} \in \mathbb{R}^{d_s}$, an appearance code $\bm z_{a} \in \mathbb{R}^{d_a}$, and a camera pose $\bm \theta \in \mathbb{R}^3$ sampled from prior distributions as input, the generator $G$ outputs an image $I$ bearing the corresponding properties:
\begin{equation}
    G:({\bm z_s}, {\bm z_a}, {\bm \theta}) \in \mathbb{R}^{d_s+d_a+3} \rightarrow I \in \mathbb{R}^{H\times W\times 3}.
\end{equation}
To achieve 3D-consistent image generation given different camera poses, we adopt radiance field as our underlying 3D representation and follow a typical 3D-aware GAN training paradigm as in~\cite{chan2021pi}. To achieve separate control of shape and appearance during image synthesis, we propose a novel generative deformable radiance field to represent these two attributes in a disentangled manner, as shown in Fig.~\ref{fig:frame}. The generative deformable radiance field consists of a template radiance field conditioned on the appearance code $\bm z_{a}$ that handles texture variations (Sec.~\ref{sec:template}), together with a deformation field and a correction field dealing with shape variations (Sec.~\ref{sec:deform}), both conditioned on the shape code $\bm z_{s}$. With the above architecture, an image can be obtained via a standard volumetric rendering process (Sec.~\ref{sec:render}). Considering that the whole framework is trained with a weak image-level adversarial loss, we further introduce a set of dedicated loss functions as the regularization of the two fields to ensure the deformation field and correction field characterize reasonable shape changes (Sec.~\ref{sec:loss}). Based on our learned generator, we develop a novel image inversion scheme that faithfully recovers the input image meanwhile maintains the inherent properties of the underlying deformable radiance field representation thus capable of high quality shape and texture editing (Sec.~\ref{sec:inversion}).


\section{Generative Deformable Radiance Field}
Our generative deformable radiance field models a category of topology-varying objects and synthesize images of new instances with a disentangled control over shape, appearance, and camera viewpoints. It consists of a template radiance field, a deformation field, and a correction field, described as follows. 

\subsection{Template Radiance Field}\label{sec:template}
The template radiance field $T$ determines the appearance of objects in a shared template volumetric space. It is represented by an MLP which takes an appearance code ${\bm z_a}$ as input, along with a 3D position $\bm x_t \in \mathbb{R}^3$ in the template space and a view direction $\bm d \in \mathbb{R}^3$, and predicts the color and density $({\bm c},\sigma_t) \in \mathbb{R}^4$ of $\bm x_t$:
\begin{equation}
    T: ({\bm z_a, \bm x_t, \bm d})\in \mathbb{R}^{d_a+6} \rightarrow ({\bm c}, \sigma_t)\in\mathbb{R}^4. \label{eq:template}
\end{equation}
The network structure of $T$ is shown in Fig.~\ref{fig:frame}. The first several layers of the network are not conditioned on the appearance code, which are used to generate a shared density field for all objects within the category as their template shape. To characterize appearance changes of the template, we inject the appearance code into the FiLM-SIREN blocks~\cite{chan2021pi} starting from an intermediate layer using modulations of frequencies $\gamma_a$ and phase shifts $\beta_a$ produced by an appearance code mapping network $M_a$ (see \cite{chan2021pi} for details).
With the above network structure, Eq.~\eqref{eq:template} can be re-written as:
\begin{align}
    T_{\sigma}: & \ \bm x_t\in \mathbb{R}^{3} \rightarrow  ({\bm f_t}, \sigma_t)\in\mathbb{R}^{d_f+1}. \label{eq:template_specific} \\
    T_{c}:& ({\bm z_a, \bm f_t, \bm d})\in \mathbb{R}^{d_a+d_f+3} \rightarrow  {\bm c}\in\mathbb{R}^3,
\end{align}
where $\bm f_t$ is the intermediate feature, and $T_{\sigma}$ and $T_{c}$ denote the two sub-nets.

\subsection{Deformation Field and Correction Field}\label{sec:deform}
The deformation field and the correction field are responsible for modeling shape variations within the class. As our underlying representation is an implicit one, we define an ``inverse" deformation field, which deforms a 3D position in the target space (i.e., the object space for image rendering) into the shared template space. In addition, since  both the deformation and correction are used to derive shape (density) changes given the shape code ${\bm z_s}$, 
we represent them via a single network $\Phi$. The network takes the shape code ${\bm z_s}$ and a 3D point ${\bm x} \in \mathbb{R}^3$ in the target space as input, and predicts a corresponding deformation vector ${\Delta \bm x} \in \mathbb{R}^3$ and a density correction $\Delta \sigma \in \mathbb{R}$: 
\begin{equation}
    \Phi:({\bm z_s, \bm x})\in \mathbb{R}^{d+3} \rightarrow ({\Delta \bm x}, \Delta \sigma)\in \mathbb{R}^4.
    \label{eq:deform_main}
\end{equation}
$\Phi$ is also an MLP with a set of FiLM-SIREN blocks \cite{chan2021pi} where the hidden layer features are modulated by frequencies $\gamma_s$ and phase shifts $\beta_s$ produced by an shape code mapping network $M_s$.
With the predicted deformation ${\Delta \bm x}$ and correction $\Delta \sigma$, the final density $\sigma \in \mathbb{R}$ and color ${\bm c} \in \mathbb{R}^3$ of the target space point ${\bm x}$ with view direction ${\bm d}$ can be obtained via:
\begin{align}
    \sigma(\bm z_s,\bm x) &= T_{\sigma}^{[\sigma_t]}\big(\bm x+\Phi^{[\Delta \bm x]}(\bm z_s, \bm x)\big) + \Phi^{[\Delta \sigma]}(\bm z_s, \bm x), \label{eq:deform_sigma}\\
    \bm c(\bm z_s,\bm z_a,\bm x,\bm{d}) &=T_{c}\big(\bm z_a, T_{\sigma}^{
    [\bm f_t]}\big(\bm x+\Phi^{[{\Delta \bm x]}}(\bm z_s, \bm x)\big), \bm{d}\big), \label{eq:deform_c}
\end{align}
where the superscripts denote the corresponding outputs of the networks when there are multiple ones. In a nutshell, the deformation field deforms a point in the target space to the template radiance field to obtain its density and color, and meanwhile the correction field modifies its density to handle topology variations that cannot be well modelled by 
deformation. 
It is worth mentioning that \cite{deng2021deformed} also combines deformation and correction, but their method only models 3D shape without considering appearance, and it uses an auto-decoder framework to learn signed distance fields for a given 3D mesh collection.
In contrast, we learn 3D deformation and correction using unstructured 2D images with generative radiance field modeling and we handle both shape and appearance.

It can be seen from Eq.~\ref{eq:deform_sigma} that our deformable radiance field architecture \emph{completely disentangles shape variations from appearance}, as the final density of a point only depends on the shape code $\bm z_s$ and is not affected by $\bm z_a$. Also note that although the final color of the point $x$ appears to also depend on the shape code $\bm z_s$ per Eq.~\eqref{eq:deform_c}, it is actually determined by the color code $\bm z_a$ and the deformed position of $\bm x$ in the template space. As a result, as long as $\Phi$ can deform the points on different shapes that are correspondences to the same position in the template space, a desired \emph{full disentanglement between shape and appearance control can be achieved}. 
In Sec.~\ref{sec:loss}, we introduce several regularizations to learn such a deformation.

\subsection{Image Rendering}\label{sec:render}
We render images of the radiance field via a standard volumetric rendering process as in ~\cite{mildenhall2020nerf,chan2021pi}. Given a camera pose ${\bm \theta}$, we cast rays in the target object space following the camera intrinsics and extrinsics and sample points along each ray using the hierarchical sampling strategy of ~\cite{mildenhall2020nerf}. For a camera ray ${\bm r}$ with sampled points $\{\bm x_i\}$ sorted from near to far, its corresponding color $C({\bm r})$ can be obtained via:
\begin{align}
C(\bm r) &= \sum_{i=1}^{n}W(\bm x_i)(1-\exp(-\sigma(\bm x_i)\delta_i))\bm c(\bm x_i,\bm d_i), \label{eq:render_c}\\
W(\bm x_i) &= \exp(-\sum_{j=1}^{i-1}\sigma(\bm x_j)\delta_j), \label{eq:render_w}
\end{align}
where $\sigma(\bm x_i)$ and $\bm c(\bm x_i,\bm d_i)$ are corresponding density and color of $\bm x_i$ obtained via Eq.~\eqref{eq:deform_sigma} and \eqref{eq:deform_c}, and $\delta_i$ is the distance between $x_i$ and its adjacent point. Note the shape and color codes are omitted here for brevity.

\subsection{Training Strategy}\label{sec:loss}
We follow a typical 3D-aware GAN's training strategy to learn our generative deformable radiance field using unstructured 2D image collection. Specifically, we sample shape code $\bm z_s$, color code $\bm z_a$, and camera pose $\bm \theta$ from prior distributions and generate corresponding images $G({\bm z_s}, {\bm z_a}, {\bm \theta})$ via Eq.~\eqref{eq:deform_sigma}--\eqref{eq:render_w}. We send the generated images as well as real images sampled from the training data to a discriminator $D$ and apply a non-saturating GAN loss with R1 regularization~\cite{karras2019style,mescheder2018training} and a pose regularization following \cite{deng2021gram}. To ensure that the deformation field derives reasonable correspondences for shape and appearance disentanglement, we further design several regularizations described below.

\begin{figure}[t]
    \centering
    \includegraphics[width=\linewidth]{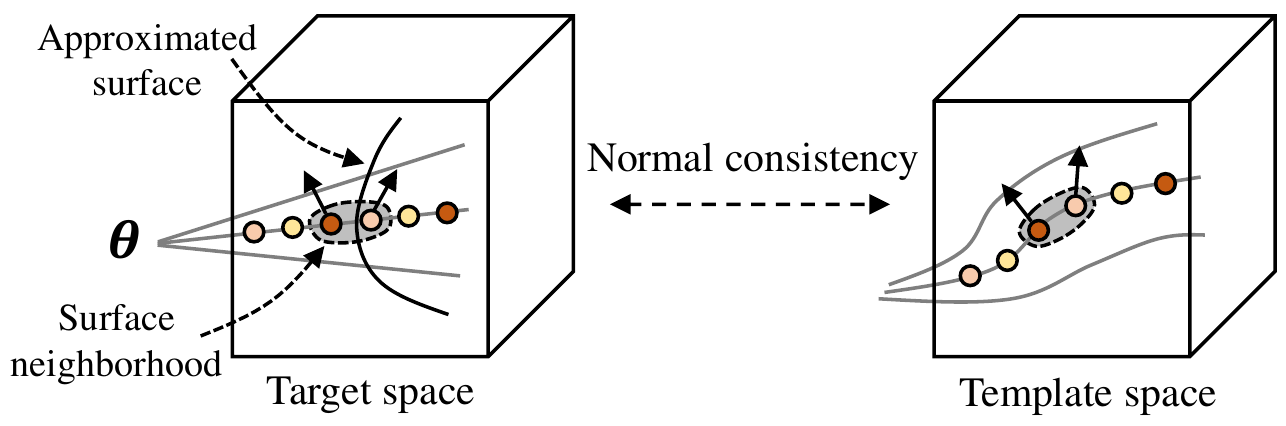}
    \vspace{-17pt}
    \caption{Illustration of the normal consistency regularization.} 
    \label{fig:normal}
    \vspace{-15pt}
\end{figure}

\noindent\textbf{Normal consistency.~} Inspired by \cite{deng2021deformed}, we enforce a normal consistency regularization between the target space radiance field and the template radiance field to ensure that the corresponding points in the two spaces derived by the deformation field have similar normal directions. For objects in a same category, their corresponding surface points with similar semantic meaning usually have similar surface normals. Therefore, the normal consistency regularization encourages the deformation field to establish correspondence between points sharing similar semantic meaning.
However, our radiance fields do not characterize well-defined object surfaces compared to the signed distance field in \cite{deng2021deformed}. To tackle this problem, we propose a new normal consistency regularization tailored for our radiance field representation (Fig.~\ref{fig:normal}):
\begin{equation}
    L_{\text{normal}} = \sum_{r=1}^{R}\sum_{\bm x_i\in\mathcal{N}(\tilde{\bm x}^{\bm r})} (1-<sg(\nabla \sigma(\bm x_i)), \nabla T_{\sigma}^{[{\sigma}]}(\bm x_i+\Delta \bm x_i)|>),
\end{equation}
where $R$ is the total number of rays. $\Delta \bm x_i$ is the deformation of point $\bm x_i$ as described in Eq.~(\ref{eq:deform_main}) which derives the correspondence between $\bm x_i$ and $\bm x_i+\Delta \bm x_i$. For $\sigma$ and $\Delta \bm x_i$, the dependency on latent code is omitted for brevity.
$\nabla$ denotes the spatial gradient of a volumetric field, which can be effectively calculated via auto-gradient of a deep learning framework such as Pytorch~\cite{paszke2019pytorch}, $sg$ denotes the \emph{stop-gradient} operator, $<\cdot,\cdot>$ is the cosine distance, and $\mathcal{N}(\tilde{\bm x}^r)$ denotes the neighborhood of the approximated surface point $\tilde{\bm x}^{\bm r}$ along ray $\bm r$. The approximated surface point can be obtained via weighted average over all points $\{\bm x_i\}$ along the ray \cite{mildenhall2020nerf}:
\begin{equation}
    \tilde{\bm x}^r = \sum_{i=1}^n W(\bm x_i)(1-\exp(-\sigma(\bm x_i)\delta_i))\bm x_i.
\end{equation}
We only apply the normal consistency loss for points near the approximated surface as the normals in the free space of a radiance field can be arbitrary. 
In practice, We adopt a hierarchical sampling strategy following \cite{mildenhall2020nerf} and only select $5\%$ of total samples that are closest to the surface point to calculate this loss.
Moreover, we do not propagate the gradient back to the final density $\sigma(\bm x_i)$ to ensure that the loss only regulates the learned deformation field and template radiance field.

\noindent\textbf{Deformation smoothness.~} We also apply a smoothness constrain on the deformation field to avoid irregular deformation:
\begin{equation}
L_{\text{smooth}} = \sum_{i=1}^n \|\nabla \Phi^{[\Delta \bm x]}(\bm x_i)\|_2.
\end{equation}
where $\|\cdot\|_2$ denotes the $L_2$ norm.

\noindent\textbf{Deformation rigidity.~} We further apply a new rigidity regularization to ensure that the learned deformation is locally rigid:
\begin{equation}
    L_{\text{rigid}} = \sum_{i=1}^{n} \|J_i^TJ_i-I\|_F,
\end{equation}
where $J_i =\nabla (\bm x_i+\Phi^{[\Delta \bm x]}(\bm x_i))$ is the Jacobian matrix of the deformed point in the template space with respect to the original point in the target space. $\|\cdot\|_F$ denotes the Frobenius norm.

\noindent\textbf{Minimal correction.~} We follow \cite{deng2021deformed} to impose a minimal correction prior  to encourage shape modeling via deformation rather than correction:
\begin{equation}
    L_{\text{correction}} = \sum_{i=1}^n|\Delta \sigma(\bm x_i)|.
\end{equation}

\begin{figure*}[t]
    \centering
    \includegraphics[width=\linewidth]{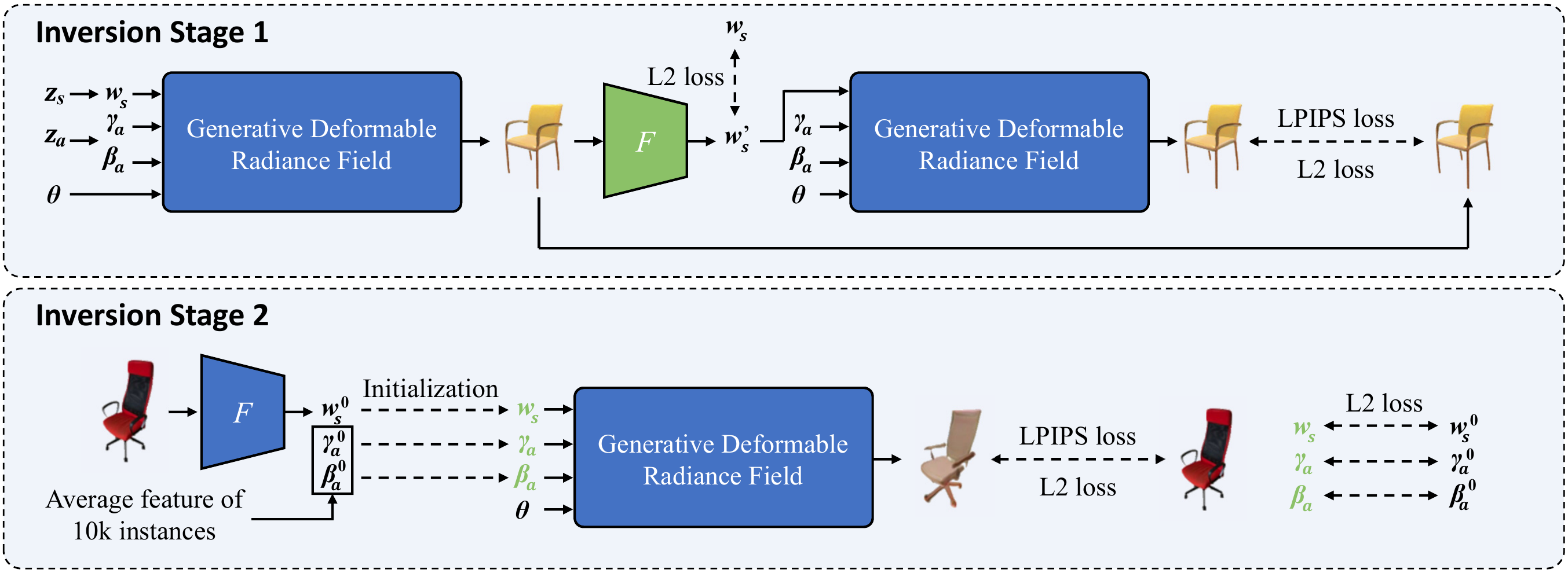}
    \vspace{-15pt}
    \caption{Overview of our image inversion pipeline. In the first stage, we train a 2D CNN using pairs of generated images and their corresponding shape and appearance features, where $\bm w_s$ denotes the last hidden feature of the shape mapping network. In the second stage, given a real image, we use the pre-trained 2D CNN to predict a shape feature as initialization. Then, we optimize the shape and appearance features to faithfully reconstruct the given image. Green color indicates the components to be optimized in each stage.} 
    \label{fig:inversion}
    \vspace{-8pt}
\end{figure*}

\section{Real Image Inversion and Editing}\label{sec:inversion}

Although our generative deformable radiance field is learned in a GAN training paradigm, it can be applied to disentangled editing of real images. To this end, we propose a novel image inversion scheme that can faithfully reconstruct the details of a given image meanwhile maintaining the shape and appearance disentanglement for image editing task. Figure~\ref{fig:inversion} shows the overview of the method.

\subsection{Inversion Strategy}
Given a real image $\hat{I}$ of an object with camera pose $\bm \theta$, we aim to find its corresponding deformable radiance field representation that can not only recovery the image content under the given viewpoint, but also support disentangled editing of shape and appearance and produce plausible results under novel views. To achieve this, a straightforward solution is to find the shape code $\bm z_s$ and color code $\bm z_a$ that describe the real image. However, obtaining such codes with high-fidelity reconstruction result for a real image is almost impossible as shown by previous GAN inversion methods~\cite{abdal2019image2stylegan,karras2020analyzing,zhou2021interpreting}. On the other hand, representing the image with intermediate features produces more faithful reconstruction result but may sacrifice the image editing quality~\cite{zhu2020domain,chan2021pi}.

Our generative deformable radiance field naturally disentangles shape and appearance for image editing. Therefore, we only need to guarantee that the recovered shape is reasonable for novel view synthesis and the recovered appearance is of high fidelity. We achieve this by optimizing the following objective function:
\begin{equation}
\begin{split}
\operatorname*{argmin }_{{\bm w_s},{\bm \gamma_a},{\bm \beta_a}}    & ~ \| I({\bm w_s},{\bm \gamma_a},{\bm \beta_a},{\bm \theta})-\hat{I}\|_2^2 + \text{LPIPS}(I({\bm w_s},{\bm \gamma_a},{\bm \beta_a},{\bm \theta}),\hat{I})\\ 
&+ \| {\bm w_s}-{\bm w_s^0} \|_2^2 + \| {\bm \gamma_a}-{\bm \gamma_a^0} \|_2^2 + \| {\bm \beta_a}-{\bm \beta_a^0} \|_2^2, \label{eq:inversion}
\end{split}
\end{equation}
where $\bm w_s$ is the last hidden feature of the shape mapping network, $\bm \gamma_a$ and $\bm \beta_a$ are output features of the appearance mapping network which are also the frequencies and phase shifts of each FiLM-SIREN block of the template field network, and $\cdot^0$ denotes the initial features for optimization (see Sec.~\ref{sec:feat_init}). $I({\bm w_s},{\bm \gamma_a},{\bm \beta_a},{\bm \theta})$ denotes the generated image and LPIPS$(\cdot,\cdot)$ is the perceptual loss from \cite{zhang2018unreasonable}.
We optimize the hidden features of the mapping network instead of the frequencies and phase shifts for shape to ensure that the recovered 3D geometry is more plausible under novel views, and we directly optimize the frequencies and phase shifts for appearance to obtain high-fidelity textures. 

\subsection{Inversion Initialization}\label{sec:feat_init}
For object categories with diverse structures such as chairs, a proper initialization of the features to be solved in the inversion optimization is important. 
The following feature initialization strategy is proposed in this work.
First, we randomly generate a collection of images $\{I^i\}$ using our trained deformable radiance field generator and record their corresponding camera poses $\{\bm \theta^i\}$ and features $\{\bm w_s^i\}$, $\{\bm \gamma_a^i\}$, and $\{\bm \beta_a^i\}$. Then, we train a CNN $F$ that predicts feature $\bm w_s$ using the following loss function:
\begin{equation}
\begin{split}
\operatorname*{argmin }_{F} &~\| F(I^i)-{\bm w_s^i} \|^2 + \| I(F(I^i),{\bm \gamma_a^i},{\bm \beta_a^i},{\bm \theta^i})-I^i \|^2 \\
& + \text{LPIPS}(I(F(I^i),{\bm \gamma_a^i},{\bm \beta_a^i},{\bm \theta^i}),I^i),
\end{split}
\end{equation}
which enforces $F$ to reproduce the shape features $\bm w_s^i$ and the image $I^i$. After training, given a real image $\hat{I}$, we predict its corresponding shape feature $F(\hat{I})$ as the initial value $\bm w_s^0$ in Eq.~\eqref{eq:inversion}.
For the appearance features, we empirically found that their initialization has minor influence on the inversion result, so we directly sample $10$K random appearance codes $\bm z_a$ and calculate their average frequencies and phase shifts as the initial values $\bm \gamma_a^0$ and $\bm \beta_a^0$ in Eq.~\eqref{eq:inversion}.

\subsection{Disentangled Image Editing}
Given a real image $\hat{I}$, we can first obtain its corresponding shape and appearance features $\bm w_s$, $\bm \gamma_a$, and $\bm \beta_a$ via Eq.~\eqref{eq:inversion}. Then, we can achieve a disentangled editing over its shape and appearance by simply replacing the features with those of another image, and synthesize multiview images of the editing results given arbitrary camera viewpoints. 
For instance, given two real images $\hat{I}^i$ and $\hat{I}^j$, we can easily obtain a new editing result $I(\bm w_s^i,\bm \gamma_a^j,\bm \beta_a^j,\bm \theta)$, which has the shape of $\hat{I}^i$ and the appearance of $\hat{I}^j$.

\section{Experiments}
\subsection{Implementation Details}
We evaluate our methods on four datasets: PhotoShape chair~\cite{photoshape2018}, CARLA~\cite{schwarz2020graf,dosovitskiy2017carla}, ShapeNet car, and ShapeNet plane~\cite{shapenet2015}.
For PhotoShape chair, we render a single view image for each chair instance with a random camera viewpoint sampled from the upper hemisphere with 30 degrees pitch variation, and obtain $15$K images in total. For CARLA, we use the $10$K rendered images provided by the dataset. For ShapeNet car and plane, we render three views for each object and obtain $10$K and $12$K total images respectively, following a similar camera distribution as in PhotoShape chair.
Our deformation network is an MLP with 4 layers with the FiLM-SIREN structure~\cite{chan2021pi}. The template network has 5 layers with sine activation for density generation, and another 4 FiLM-SIREN layers for color output.  
Our discriminator is adapted from \cite{chan2021pi}, and the inversion initialization CNN is a ResNet-34 network. 
During training, we sample shape code $\bm z_s$ and appearance code $\bm z_a$ from a normal distribution of dimension $256$. We use a single random code for both shape and appearance of a generated instance. We randomly sample camera pose $\bm \theta$ from the same distribution used to generate training data. 
We use Adam optimizer and train the model progressively from $64\times64$ resolution to $128\times128$ with batchsize 32 on 8 Nvidia Tesla V100 GPUs with 32GB memory. The training takes 3 to 6 days.

\begin{figure*}
    \centering
    \includegraphics[width=\linewidth]{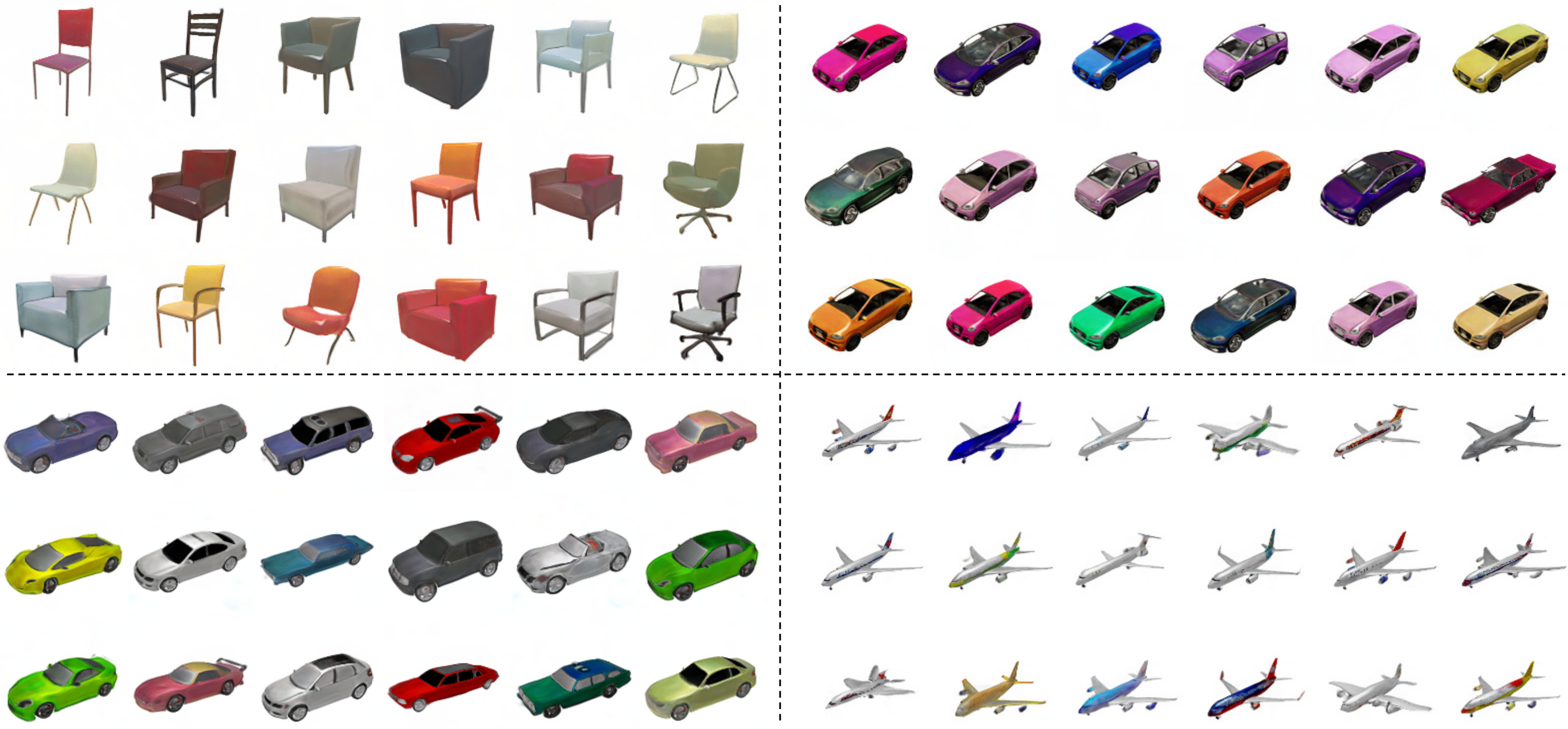}
    \caption{Random image generation results of our method on four different datasets. (\textbf{Best viewed with zoom-in})}
    \label{fig:gen_results}
    \vspace{-1pt}
\end{figure*}

\begin{figure*}[t]
    \centering
    \includegraphics[width=\textwidth]{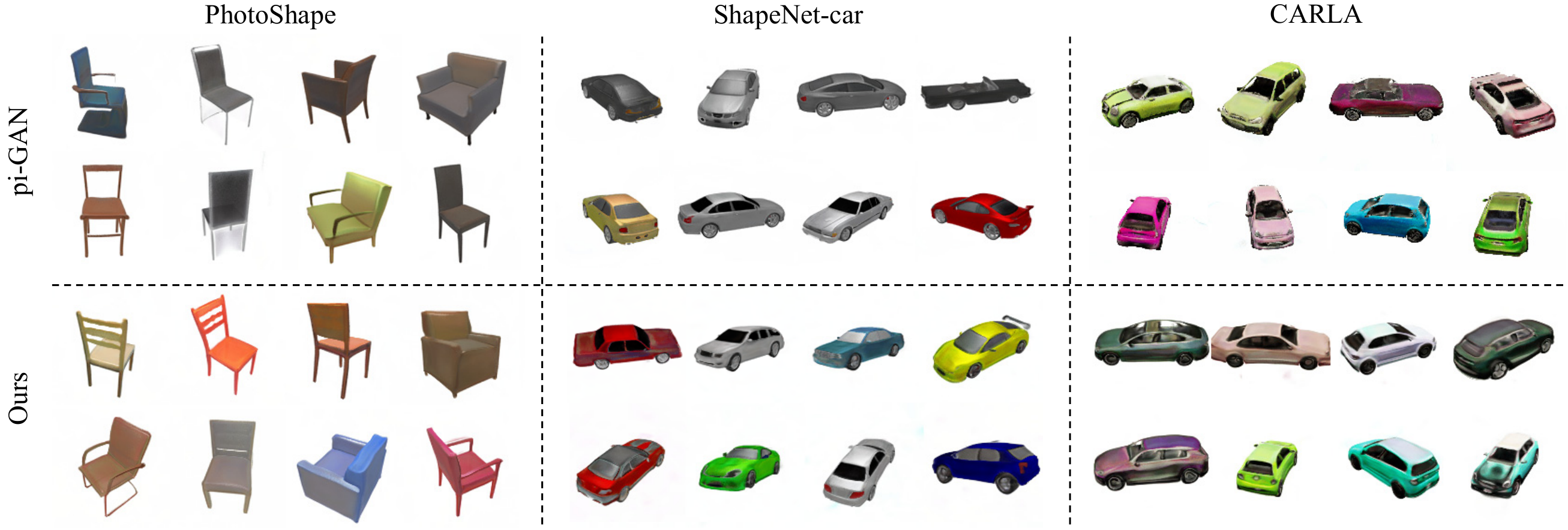}
    \caption{Qualitative comparison with the 3D-aware GAN model pi-GAN. Note that pi-GAN does NOT address shape-appearance disentanglement as we do, and that our method is trained on the same data without extra labels for disentanglement.}
    \label{fig:compare_pigan}
    \vspace{-2pt}
\end{figure*}

\subsection{Random Generation Results}

Figure~\ref{fig:gen_results} shows the random image generation results by our method on four different datasets. Our method can generate images of various objects with diverse structures and freely change camera viewpoints. The multiview generation results can be found in the accompanying video.


\subsection{Disentangled Generation Results}
\begin{figure*}[t]
    \centering
    \includegraphics[width=\textwidth]{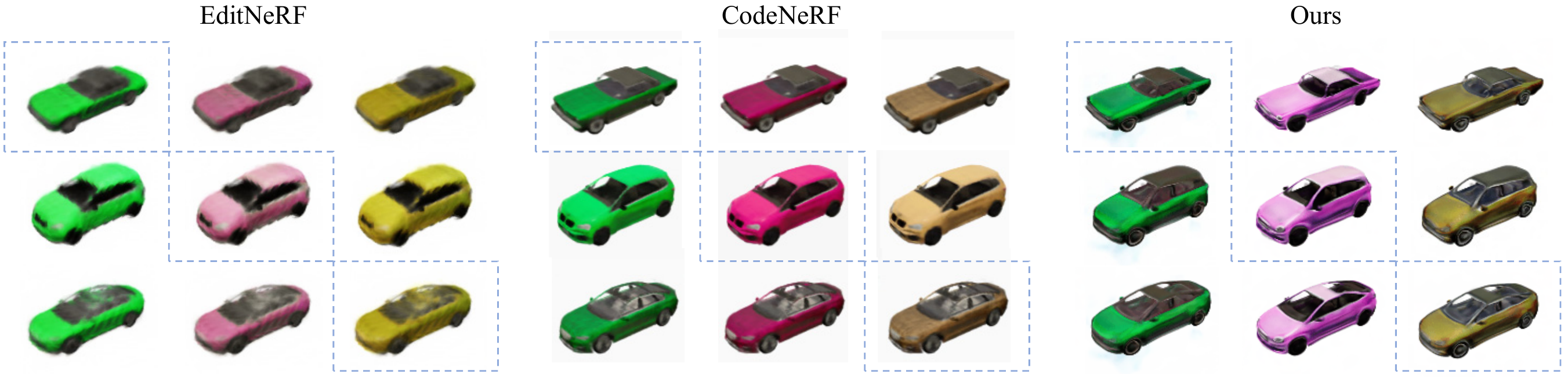}
    \vspace{-14pt}
    \caption{Qualitative comparison of code swapping results on CARLA. The images on the diagonal are the reference instances and others are generated by combining the row-wise shape code and column-wise appearance code.
    Our method achieves much better shape and appearance disentanglement than EditNeRF and CodeNeRF (e.g., see the color of car roof in the column-wise texture swapping results) }
    \label{fig:compare_swap}
    \vspace{-5pt}
\end{figure*}

Our method enables a disentangled control over shape and appearance attributes of the generated images. We achieve this by varying either the shape code or appearance code and keeping the other unchanged. The corresponding generation results are shown in Fig.~\ref{fig:teaser} and Fig.~\ref{fig:compare_swap} (see video for multiview results).
With our generative deformable radiance field and the carefully designed training strategy, we obtain plausible dense correspondence between different objects. Textures can be correctly transferred between shape structures of similar semantic meaning. Note that our method is only trained with  2D images without any semantic labels.

\begin{figure}[t]
    \centering
    \includegraphics[width=\linewidth]{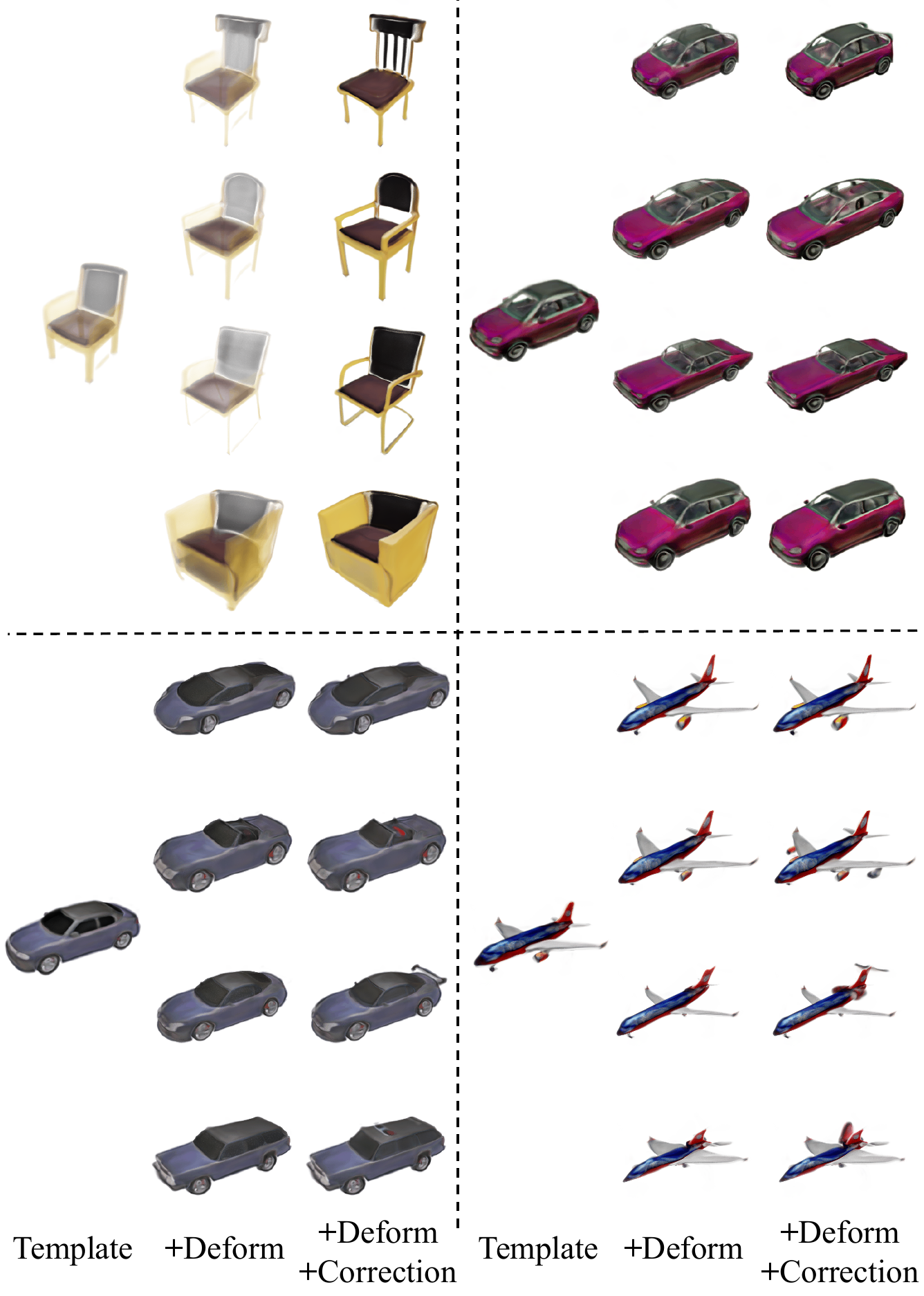}
    \vspace{-15pt}
    \caption{The learned templates and generated shapes by our method. (\textbf{Best viewed with zoom-in})}
    \vspace{-13pt}
    \label{fig:template}
\end{figure}

In Fig.~\ref{fig:template}, we illustrate the the learned template fields for different categories and how different shapes are generated through deformation and correction.

\begin{table}[t]
\caption{Quantitative comparison with pi-GAN.}
\begin{tabular}{l|ll|ll|ll}
\centering
\multirow{2}{*}{Methods} & \multicolumn{2}{l|}{\!\!PhotoShape 128\!\!} & \multicolumn{2}{l|}{\!\!ShapeNet-car 128\!\!} & \multicolumn{2}{l}{\!\!CARLA 128\!\!} \\
                         & FID          & KID          & FID         & KID         & FID         & KID         \\ \hline
pi-GAN                   & 19.4         & 0.71         & 18.9        & 0.81        & 41.5        & 2.10        \\
Ours                     & 23.8         & 1.09         & 27.7        & 1.54        & 41.2        & 2.07       
\end{tabular}
\vspace{-5pt}
\label{table:fid_pigan}
\end{table}

\begin{figure}[t]
    \centering
    \includegraphics[width=\linewidth]{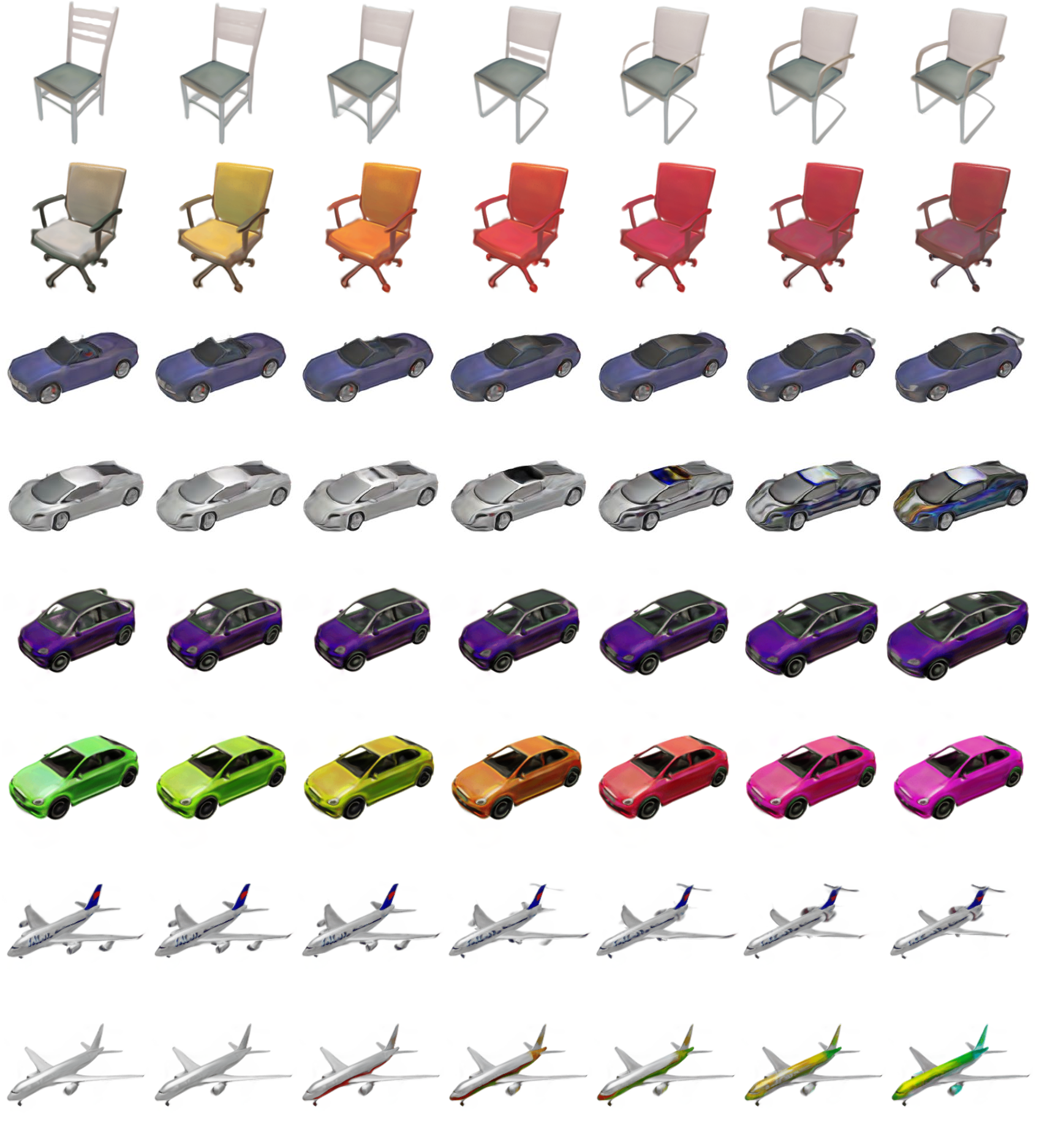}
    \vspace{-15pt}
    \caption{Latent space interpolation of our method on four different datasets. Shape and appearance interpolations are in odd and even rows, respectively.  (\textbf{Best viewed with zoom-in})}
    \vspace{-10pt}
    \label{fig:latent_interp}
\end{figure}

We further show the  latent space interpolation results in Fig.~\ref{fig:latent_interp}. Our disentanglement design makes it possible to interpolate either shape and appearance while keeping the other unchanged. The interpolation produces reasonable intermediate results for two shapes with topology difference.

\subsection{Comparison with Prior Art}
\textbf{Comparison with pi-GAN.~} First, we compare our uncurated generation results with pi-GAN~\cite{chan2021pi}, and
Fig.~\ref{fig:compare_pigan} shows the visual results. Our method achieves competitive generation quality, but further enables shape and appearance disentanglement which cannot be achieved by pi-GAN. Table~\ref{table:fid_pigan} further shows the quantitative comparison, where we evaluate the FID~\cite{heusel2017gans} and KID~\cite{binkowski2018demystifying} scores using $5$K real images and $5$K generated images. Compared to pi-GAN, our method achieves disentangled control of shape and appearance with only a slight increase of FID and KID.

\begin{figure*}[t]
    \centering
    \includegraphics[width=\linewidth]{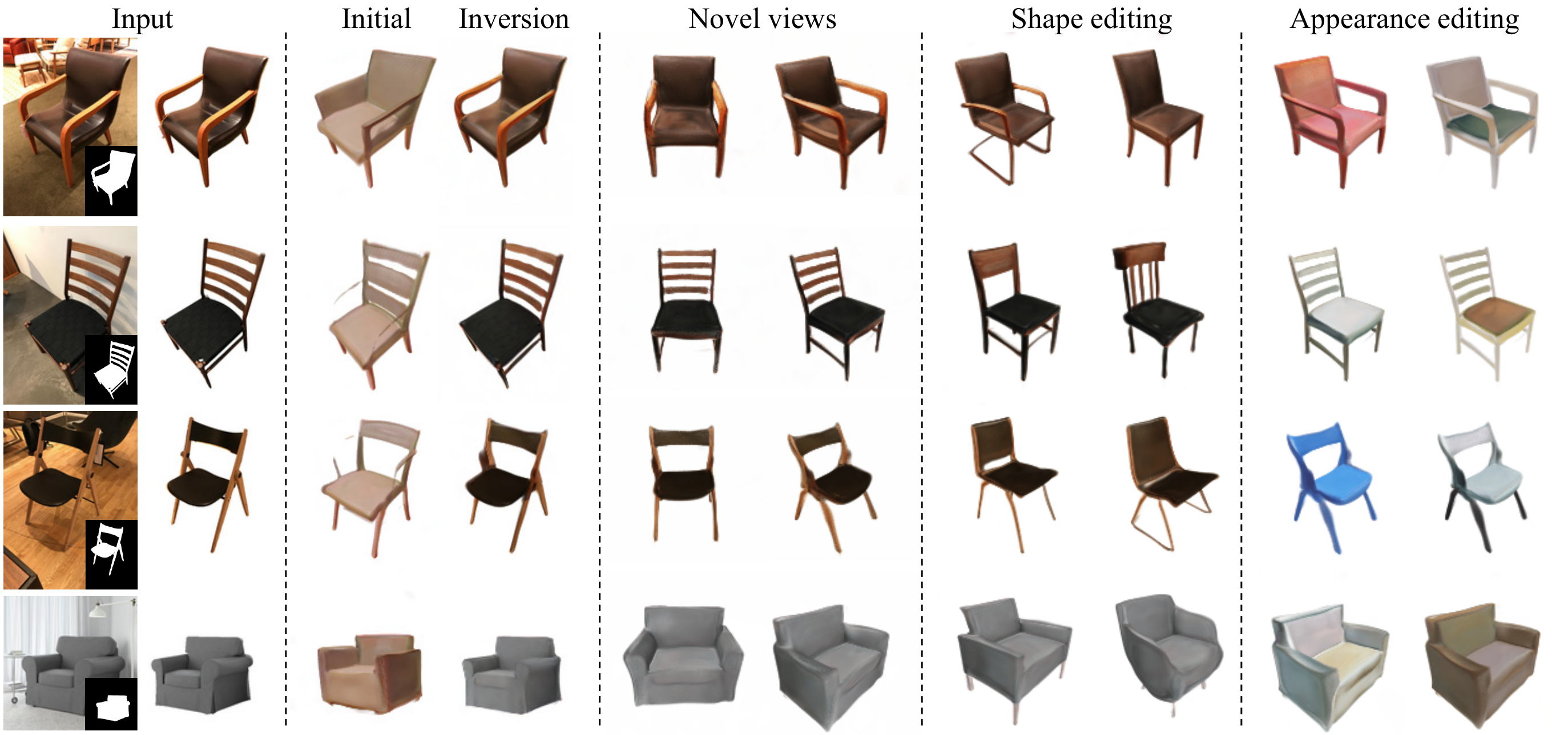}
    \caption{Real image inversion and disentangled editing results from our method.}
    \label{fig:inv_results}
    \vspace{-10pt}
\end{figure*}

\noindent\textbf{Comparison with CodeNeRF and EditNeRF.~}
To further validate the disentangled image synthesis ability of our method and compare with the state-of-the-art methods, we conduct shape and appearance code swapping experiments and compare with CodeNeRF~\cite{jang2021codenerf} and EditNeRF~\cite{liu2021editing}, both of which can separately control the shape and appearance attributes of an instance. 
Figure~\ref{fig:compare_swap} shows the code swapping results of the generated CARLA objects. CodeNeRF and EditNeRF can hardly maintain a correct texture transfer result when swapping the appearance code between different instances (e.g., see the roof of the cars). By contrast, our method well keeps the texture consistency between corresponding semantic parts of different objects.



\begin{table}[t]
\centering
\caption{Ablation study on different network components and training losses on PhotoShape-chair $64\times64$.}
\begin{tabular}{l|cc}
Setting                  & FID            & KID ($\times100$)               \\ 
\hline
no deformation~ ~ ~~          & 24.6   & 1.28                                             \\
no correction~ ~ ~            & 18.4    & 0.59                                              \\
w/o deformation smooth. loss~ ~ ~~  & 17.7    & 0.70                                            \\
w/o rigidity loss~ ~ ~~       & 15.8    & \textbf{0.56}                                        \\
w/o normal consist. loss~ ~ ~~      & 18.6    & 0.82                                             \\
w/o minimal correction loss~ ~ ~~  & 16.0   & 0.61                                   \\
\hline
Full model                & \textbf{15.6} & \textbf{0.56}                           
\end{tabular}
\vspace{-6pt}
\label{table:ablation}
\end{table}

\subsection{Real Image Inversion and Editing}
Figure~\ref{fig:inv_results} shows some typical real image inversion and editing results by our method.  
Since our model is trained without background, we remove the irrelevant  background content before inversion. 
As shown, our method can faithfully reconstruct the shape and appearance of the given real images and render plausible novel views of them, even though it is only trained with synthetic data. In addition, we can further edit the shape and appearance of the real images by replacing the inverted shape or appearance features with randomly generated ones. 






\begin{figure}
    \centering
    \includegraphics[width=\linewidth]{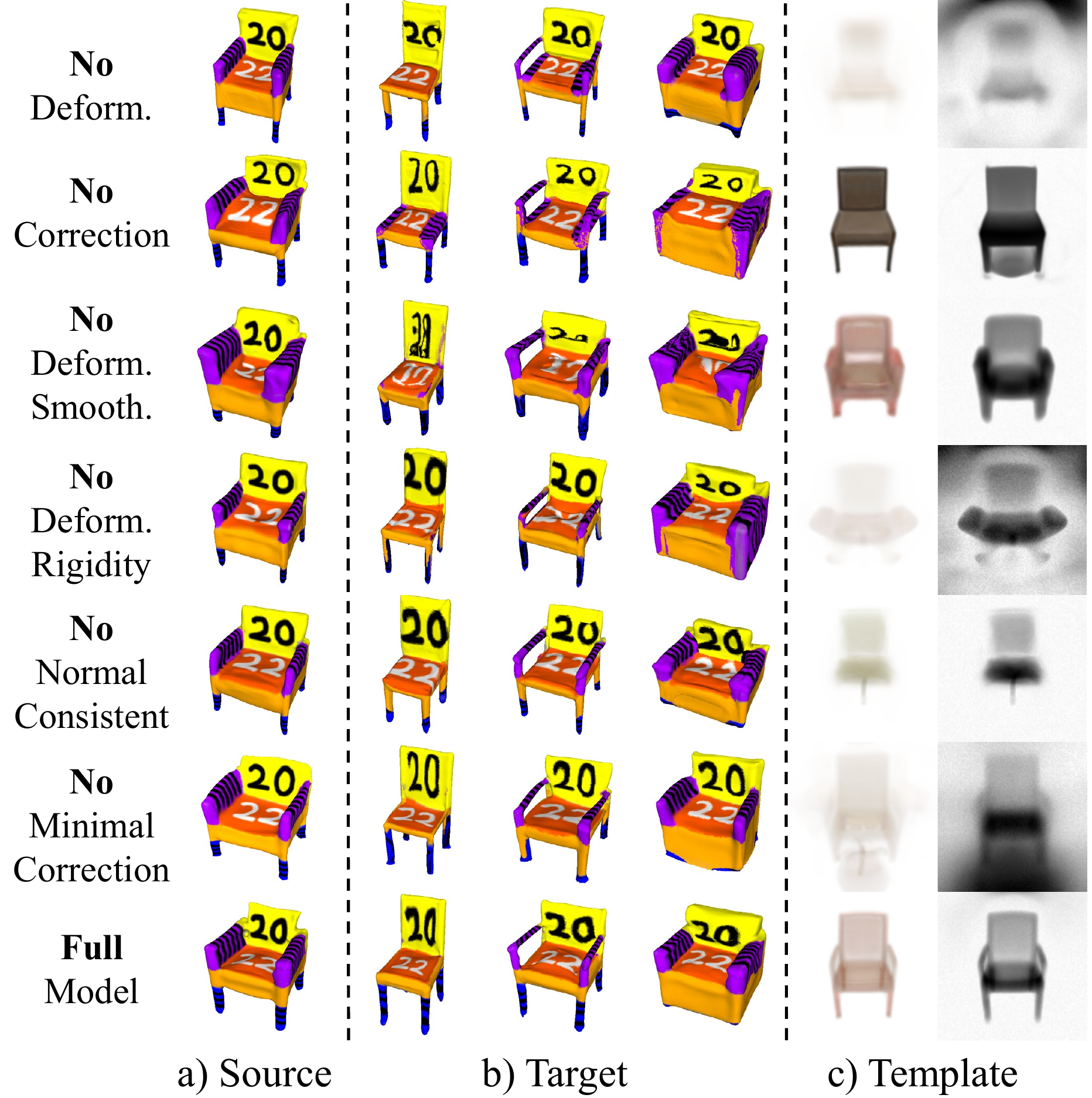}
    \vspace{-13pt}
    \caption{Visualization of the learned dense correspondence and template radiance field under different setups.}
    \label{fig:corre_vis}
    \vspace{-5pt}
\end{figure}

\begin{figure*}
    \centering
    \includegraphics[width=\linewidth]{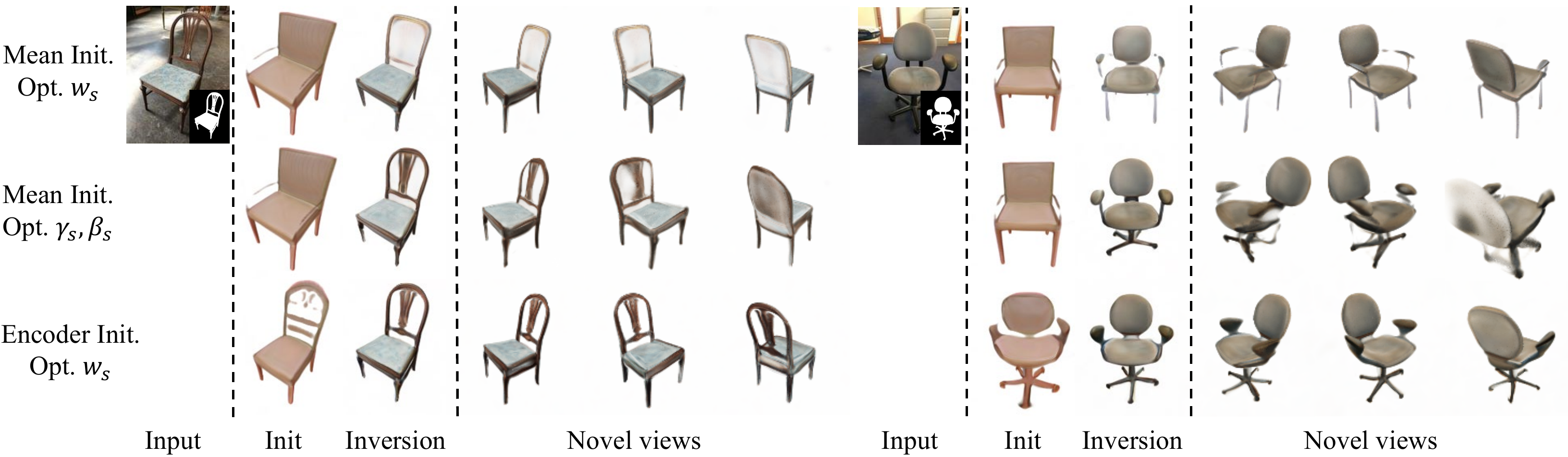}
    \vspace{-14pt}
    \caption{Comparison of different inversion initialization methods.}
    \label{fig:inversion_compare_v1}
    \vspace{-5pt}
\end{figure*}

\subsection{Ablation Study}
We evaluate the efficacy of our deformable radiance field architecture and training regularization on PhotoShape chairs with $64\times64$ resolution. Table~\ref{table:ablation} shows the influence of different components on image generation quality. Our full model with deformation field, correction field, and all regularization losses yields the best result. 

We further evaluate the influence of each component on the learned dense correspondence between different shapes in Fig.~\ref{fig:corre_vis}. To visualize dense correspondence, we extract isosurfaces from the density fields via MarchingCubes~\cite{lorensen1987marching} and manually painted some textures on the extracted meshs, as shown in Fig.~\ref{fig:corre_vis}(a). With a textured mesh as source shape, we can transfer its appearance to other generated instances using the dense correspondence derived from the learned deformation field, and Fig.~\ref{fig:corre_vis}(b) shows the texture transfer results. 
We also visualize the learned template radiance fields in Fig.~\ref{fig:corre_vis}(c) where we render the corresponding color and depth images. 

As shown in in Fig.~\ref{fig:corre_vis}, the correspondence is poor without the deformation or correction. The model without deformation can only derive correspondence by absolute spatial location, and without correction it is unable to separate parts with topology difference. The deformation smoothness and rigidity losses alleviate irregular deformation and local distortions. 
The normal consistency loss encourages the deformation field to learn a more precise correspondence, and the minimal correction loss prevents the density from being compensated mainly by the correction field and hence also leads to improved correspondence. Our full model ends up with a semantically informative template and yields the best quality in terms of dense correspondence.

To further verify the effectiveness of our proposed CNN-based inversion initialization, we compare it with the commonly-used mean latent code initialization strategy. Specifically, we first obtain a mean shape latent feature $\bm w_s^0$ by averaging the features of 10K randomly generated instances. Then we test two settings: directly optimizing $\bm w_s$ with $\bm w_s^0$ as initialization (i.e., same as ours defined in Eq.~\ref{eq:inversion}), or optimizing the latent features $(\bm \gamma_{s}, \bm \beta_{s})$ with the corresponding values derived from $\bm w_s^0$ as initialization. As shown in  Fig.~\ref{fig:inversion_compare_v1}, the former strategy leads to poor image reconstruct quality with inaccurate geometry and appearance. The latter one can well reconstruct the given image, but the geometry is erroneous and unnatural as can be seen from the novel view renderings. In contrast, our CNN-based initialization scheme can estimate a good starting point  for the inversion process in the $\bm w_s$ space, which leads to our high-quality reconstruction and novel view rendering results.

\subsection{Running time}
It takes our method 0.4 seconds to render a 128$\times$128 image on a Nvidia Tesla V100 GPU. For image inversion, we use 2K optimization steps which takes about 15 minutes for a single instance. For our image editing experiments, we simply replace the shape or appearance features and hence they take the same amount of time as image rendering.

\subsection{Failure Cases}
\begin{figure}
    \centering
    \includegraphics[width=1.0\linewidth]{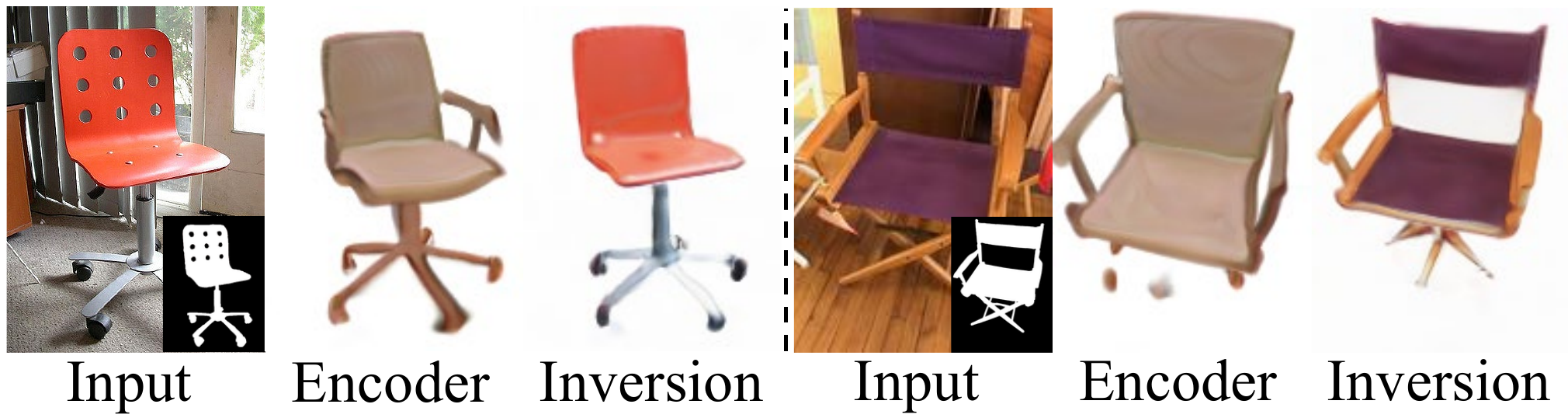}
    \caption{Some failure cases of real image inversion from our trained model.}
    \label{fig:fail_case}
    \vspace{-3pt}
\end{figure}

We illustrate two failure cases of image inversion in Fig.~\ref{fig:fail_case}. Our method may not be able to recover the correct geometry when the shapes in the input images are rare or the local structures are too complex. These cases are difficult to handle for both our trained generator and the inversion initialization CNN.


\section{Conclusion}
We have presented a novel generative deformable radiance field that can generate multiview images of an object category with topology variations, and achieve disentangled control over shape and appearance attributes of the synthesized images. 
The shape-appearance disentanglement is achieved in an unsupervised manner through our deliberate architecture and training loss design and no additional label is introduced compared to previous 3D-aware GAN training.
We also demonstrated that our method can be applied to disentangled editing of real images by leveraging our proposed image inversion scheme. We believe that our method has strong potentials for realistic
virtual content creations and manipulations in future applications.

\bibliographystyle{eg-alpha-doi}  
\bibliography{ref} 


\end{document}